\documentclass[utf8]{FrontiersinHarvard} 

\usepackage{url,hyperref,lineno,microtype,subcaption}
\usepackage[onehalfspacing]{setspace}
\usepackage[normalem]{ulem} 
\usepackage[]{inputenc}
\usepackage[english]{babel}
\usepackage{longtable}
\usepackage{array}
\usepackage{ragged2e}
\usepackage{booktabs}
\usepackage{url}

\renewcommand{\cite}{\citep}

\def\keyFont{\fontsize{8}{11}\helveticabold }
\def\firstAuthorLast{Đuranović \& Robnik-Šikonja} 
\def\Authors{Vuk Đuranović\,$^{1}$, Marko Robnik Šikonja\,$^{1*}$}
\def\Address{$^{1}$Faculty of Computer and Information Science, University of Ljubljana, Slovenia }
\def\corrAuthor{Marko Robnik Šikonja}
\def\corrEmail{marko.robnik@fri.uni-lj.si}

\begin{document}
\onecolumn
\firstpage{1}

\title[ Query-focused summarization for less-resourced languages]{QFS-Composer: Query-focused summarization pipeline for less resourced languages} 

\author[\firstAuthorLast ]{\Authors} 
\address{} 
\correspondance{} 

\extraAuth{}

\maketitle

\begin{abstract}

\section{}
Large language models (LLMs) demonstrate strong performance in text summarization, yet their effectiveness drops significantly across languages with restricted training resources. This work addresses the challenge of query-focused summarization  (QFS) in less-resourced languages, where labeled datasets and evaluation tools are limited. We present a novel QFS framework, QFS-Composer, that integrates query decomposition, question generation (QG), question answering (QA), and abstractive summarization to improve the factual alignment of a summary with user intent. We test our approach on the Slovenian language. To enable high-quality supervision and evaluation, we develop the Slovenian QA and QG models based on a Slovene LLM and adapt evaluation approaches for reference-free summary evaluation. Empirical evaluation shows that the QA-guided summarization pipeline yields improved consistency and relevance over baseline LLMs. Our work establishes an extensible methodology for advancing QFS in less-resourced languages.

\tiny
 \keyFont{ \section{Keywords:} summarization, large language models, less-resourced languages, question answering based evaluation, Slovene} 
\end{abstract}

\section{Introduction}

In recent years, the world has witnessed rapid progress of generative artificial intelligence (AI) models capable of producing logical and complete units of text. Models like ChatGPT \citep{chatgpt} and Llama \citep{llama}, which were trained on large amounts of data collected from the internet, have changed the way we work through their ability to respond to queries or prompts from human users. Although breakthroughs and improvements in large language models (LLMs) are made daily, the research community has noted differences in performance across languages. For example, the recent study by \citet{high_vs_low} showed that LLMs perform better in machine translation for languages that were highly represented in the training set, such as English, Chinese, and Spanish, compared to less-resourced languages.   

We aim to address this performance gap in query-focused summarization (QFS), also known as target-based summarization, a task in which the goal is to summarize a given text with respect to the query provided to the model. A common use case for QFS is in public relations and media press clipping, where companies are interested in how their firm is portrayed in a given text or how news may affect their business. 

The crucial in our work is performance evaluation. For the summarization task, a common practice is to compare the LLM's summary with a reference summary created by qualified human experts. Although this approach has its benefits when we want to train the model to respond in a certain style, the downside is that collecting reference summaries is a time-consuming and expensive process. Therefore, following other works on less-resourced languages, we resort to automatic reference-less evaluation metrics \cite{qags, questeval, rquge}. However, most of these evaluation metrics were developed for English, and we need to adapt them for our language of choice.

At a high level, contributions of this work are the following :
\begin{itemize}
    \item Our own QFS system called \textit{\textbf{QFS-Composer}} that is based on question answering\footnote{The framework is available at GitHub repository: \url{https://github.com/VukMNE/qfs-composer-paper}}.
    \item Working QG\footnote{QG model available at: \url{https://huggingface.co/VukDju/GaMS-9B-Instruct-QG-Full-3ep}}and QA\footnote{QA model available at: \url{https://huggingface.co/VukDju/GaMS-9B-Instruct-QA-3ep}}
    \item Translation of MOCHA dataset\cite{mocha} to the Slovenian language \footnote{Dataset available at: \url{https://github.com/VukMNE/MOCHA_slo}} 
    \item Adaptation of QuestEval\footnote{QuestEval for the Slovenian language \url{https://github.com/VukMNE/QuestEval}} QAGS\footnote{Implementation is available inside the QFS-Composer repository, see footnote 1.}, RQUGE\footnote{RQUGE for the Slovenian language:\url{https://github.com/VukMNE/RQUGE}} for evaluation of summaries in Slovene and comparison of summarization systems.
    \end{itemize}

The remainder of this paper is organized as follows: In section 2, we present the related work. The third section introduces the QFS-Composer, the framework for target-based summarization and it's components, as well as instructions for their development. In section 4, we evaluate our approach against baseline LLMs, and present results. The section 5 contains discussions about results, limitations of the approach and possible next steps in the research.

\section{Related work}
Our outline of related works starts with developments in extractive and abstraction summarization, and addresses also evaluation and components of QFS. We also shortly cover existing efforts in summarization for Slovene.

With the increasing volume of text data published on the internet, the focus of many studies has been on finding efficient mechanisms to summarize extensive text documents. A comprehensive survey providing insights into recent works on this topic is provided by  \citet{el_kassas_comprehensive_survey}. 
Based on the approach, summarization methods are classified as either extractive or abstractive. Extractive methods rank sentences within the text based on their probability of being included in the summary, and then compose the summary by concatenating the top-ranked sentences into the output summary, without modifying them. On the other hand, abstractive methods focus on paraphrasing the main points of the observed document. 

Early extractive approaches relied on sentence-sentence similarity scores to select sentences for inclusion in a summary. Among these, Maximal Marginal Relevance (MMR) \cite{mmr}, achieved the best results by balancing each sentence’s relevance with the novelty it adds relative to the sentences already in the summary. Subsequent work explored graph-based methods such as TextRank \cite{textrank}, which represents sentences as vertices in a similarity graph and applies PageRank algorithm \cite{pagerank} to select the most central ones. Another well-known example of a graph-based extractive text summarization technique is LexRank \cite{lexrank}, whose similarity measure uses TF-IDF-weighted cosine similarity, which accounts for term importance across documents. The graph construction in LexRank introduces thresholds for edge creation between sentences resulting in a sparser graph. A detailed overview of extractive methods is provided in a survey by \citet{extractive}.

Abstractive methods must not only identify the main points of the text, but also generate new wording based on those points. Consequently, they require more sophisticated architectures, such as neural networks with an encoder–decoder architecture. Abstractive summarization has evolved from RNN-based sequence-to-sequence models that used attention mechanisms like the one presented by \citet{rush2015neural}, and has since shifted towards large-scale pre-trained transformers such as BART\cite{bart}, PEGASUS \cite{pegasus}, and GPT variants \cite{chatgpt}. These LLMs produce smooth summaries that capture the meaning of the original text but rephrasing it. Some abstractive models suffer from the lack of named entity inclusion due to the design that requires them to paraphrase content, but the work by \citet{ne_inclusion} provides a way to overcome this problem. A comprehensive study of recent abstractive methods is provided by \citet{abstractive}. 

The abstractive methods have to assure factual consistency of the summary with the original text, as LLM hallucinations can introduce information that is not present in the observed document. While extractive methods ensure the generated summary contains salient parts of the source text, this approach may produce incoherent output. A hybrid approach called extractive-then-abstractive summarization \cite{liu2021combined, tretyak2020combinationabstractiveextractiveapproaches} first extracts relevant sentences from the text using standard extractive methods and forwards them to an LLM to re-formulate them into a coherent summary. 

The extractive-abstractive approaches exists also in the field of QFS. The early approaches \cite{qfs_ext1, qfs_ext2} were based on extractive methods that compared the similarity of the texts' salient segments with the query, using a similarity score. Recently, abstractive transformer-based methods have offered better comprehensiveness of the text. \citet{ltr} implemented LTR (Learning To Rank) algorithm on a transformer architecture where decoder is shared by the summarization and LTR task. By forcing the same decoder to do both summarization and ranking of overlapping fixed-length segments, gradients from the ranking loss function push the decoder’s attention to focus on parts of the source document that belong to high-ranked segments. A graph-based approach GraphRAG \cite{graphRag} harnesses the power of LLMs to extract a knowledge-graph from a large corpus of text documents. The graph is partitioned hierarchically into nested communities of related entities, and for every community the LLM produces a short report describing its entities, relations, and salient claims. This process, called "Indexing", is performed only once during the initial analysis of text corpora and when new data are inserted. When a user forwards a query to this system, the LLM reads community summaries generated in the indexing phase and reasons over them in order to create a response to the query. GraphRAG has a high commercial value because it allows businesses to utilize LLMs inside their organizations without exposing their private data to third parties.

Query-focused summarization is highly overlapped with the question-answering (QA) task, sometimes referred to also as factoid text summarization. These question answering models are trained to output simple text units, often a group of words from the original text that are a correct answer to a question. A complementary task to QA is question generation (QG), where the model generates questions based on the observed document. A common approach is to pass the text to a named entity recognition (NER) model, which then detects and extracts named entities present in the text, and then those named entities are fed to a QG model that creates questions using them.

Both QG and QA are used in our QFS approach as building blocks of summarization models and for the evaluation of summaries. \citet{summeval} provides a thorough analysis of the correlation between human quality judgments and automatic reference-based evaluation metrics for summarization, such as ROUGE \cite{rouge} (and its many variations),  BERTscore \cite{bertscore}, BLEU \cite{bleu}, etc. According to the authors, QAGS \cite{qags}, a QA-based evaluation methodology, showed a stronger correlation with human preference than standard evaluation scores. QuestEval \cite{questeval}, another approach to evaluation based on question generation and answering, which builds upon the QAGS framework, also showcased superior performance compared to ROUGE and other standard metrics based on comparison with the reference summaries. 

As high-quality summary should provide answers to multiple subquestions expressed in the users' query, the focus of scientific research has been on decomposing the query into multiple questions or steps, expressed in natural language, that the AI model should solve sequentially. \citet{qdmr} introduced a Question Decomposition Meaning Representation (QDMR), a supervised approach for query decomposition by training a model that can parse their decomposed representation from the complex question. The idea behind QDMR is that AI models should break down complex queries into simple steps, where each step requires the model to find an answer in the source documents or within the text corpora it is trained on. The LLM should solve these QDMR subtasks sequentially, and answers to earlier questions are leveraged to solve later ones.

For automatic text summarization in Slovene, \citet{vzagar2022cross} developed a cross-lingual transfer approach by aligning Slovenian embeddings with those of resource-rich English models. Other works adapted popular transformer-based models for summarization in Slovenian, e.g., SloT5 model \cite{t5_sl}, architecturally based on Google's T5 \cite{t5}. Recently, generative Slovene LLMs from the GaMS family (Generative models for Slovene) \cite{gams1}, based on Gemma \cite{gemma}, achieved state-of-the-art results on the SloBench benchmarking platform. Finally, the \textit{SloMetaSumm} \cite{slometasumm} system recommends the most suitable summarization model for a given text in the Slovene language. Development of these models was supported by datasets extracted from large text corpora in the Slovenian language, such as KAS \cite{kas}. Regarding the work in QFS, to the best of our knowledge, the only work is still preneural \cite{canhasi2014weighted}.

\section{Materials and Methods}
\subsection{QFS-Composer}
In this section we present the main components of our QFS system, named QFS-Composer. We start with the Query Decomposer, followed by a QA model paired with a span scorer, and end with an LLM that produces the final output. The main idea behind this architecture is to augment the original prompt and enforce factual consistency in model outputs.

\subsubsection{Query Decomposer}
\label{sec:QueryDecomposer}
Queries used in QFS can be broad, topic- or aspect-oriented, and sometimes more similar to a command or a request for information retrieval than to a real question. Moreover, a large portion of queries requires a response to multiple different subtasks or sub-questions. A frequent request in QFS is a comparison, or a summarization of a document from the perspective of an entity mentioned in the text. Hence, we believe that the best way to provide a summary that satisfies all requests from the user's query is to split the initial query into separate smaller tasks. After successfully decomposing the query, we forward the questions to the QA model to find answers in the source text, and pass the user's query, along with extracted question-answer pairs, as an augmented prompt to the model, which will then generate a summary. 

Ideally, we would train a separate supervised model for query decomposition using a dataset specifically dedicated to that purpose, as in the QDMR system \cite{qdmr}. Unfortunately, the QDMR query decomposition model was trained on the \texttt{BREAK} dataset, designed for multi-document setting that does not align with the focus of our research. Since it is not expected that such datasets would exist for less-resourced languages (we could not identify any such dataset for Slovene), we propose two possible workarounds. 

The first approach (called \textbf{\texttt{decomp\_aug}} in the experiments in Section \ref{sec:experiments}) is to utilize an LLM with a specially designed few-shot prompt to decompose the query into multiple sub-questions. A possible downside of this approach is that the LLM could introduce questions that are very different from the type of questions present in the SQuAD dataset (used to train our QA component), which could consequently lead to QA model not being able to answer them. Nevertheless, we tried this approach and the prompt we designed for an LLM to decompose the user query is available in Appendix A. 

The second possible approach (called \textbf{\texttt{ner\_aug}} in the experiments in Section \ref{sec:experiments})) suitable for less-resourced languages is to employ a named entity recognition (NER) model that scans the user’s query and extracts entities for which our QG model subsequently generates questions. The downside of this approach is its limitation to named entities as the focus of summarization. As real world applications of QFS mostly use named entities, this is an acceptable limitation. In Section \ref{sec:evaluation}, we test both approaches and report the results.

\subsubsection {QA component}
An integral part of our summarization system is a question answering model. The purpose of this component is to emphasize the factual information from the source text that is relevant to the user's query. QA component is utilized to answer the questions produced in the query decomposition, and the answers with the accompanying questions are merged with the user's query and source text in the augmented prompt that is sent to the final answer generating LLM.

As the base for our Slovene QA model, we utilized Slovene GaMS-9B-Instruct LLM, and fine-tuned it for three epochs over the Slovenian translation\cite{squad_slo} of SQuAD v2 dataset \cite{squad2_original}  The token-level cross-entropy loss function was optimized during training. The initial learning rate $\alpha$ was set to $ 2 \times 10^{-6}$. The learning rate schedule employed a linear warmup for 10\% of the total training steps, followed by linear decay. We used per-device batch size 1 with 16× gradient accumulation (effective batch size 16), AdamW (8-bit) as implemented in \texttt{adamw\_bnb\_8bit} (with weight decay set to $0.01$ and $\beta_1 = 0.9$, $\beta_2 = 0.999$ and $\epsilon = 10^{-8}$), and gradient-norm clipping set to 0.1. Training ran in \textbf{BF16} mixed precision, and the validation set evaluation was performed at the end of each epoch. The QA model was trained under the same computational setup, using a single NVIDIA A100 GPU (80 GB memory).

The integration of the QA model into our architecture is not straightforward due to limitations from the training process. In particular, due to the memory constraints on our hardware, we cap the QA's model input length at 350 characters. Preliminary experiments with longer inputs resulted in frequent out-of-memory errors, so 350 characters was chosen as the maximum stable value across runs. This restriction means that the model without adjustments would not be able to answer a question when given a longer document as context, especially if the answer lies in the middle or the end of the document.

To mitigate this issue, we split longer textual content into smaller chunks that the QA model could process. For each chunk, we compute BERTScore with respect to the question and sort the chunks in descending order by this score. We let the QA model answer only $n$ top-ranked chunks for which the BERTScore is higher than threshold value of $0.85$, and we take the modal prediction as final answer to the question. If there is a tie, the answer from the chunk with a higher BERTScore is selected as the output. The BERTScore threshold of 0.85 was selected heuristically during the development phase based on manual inspection of several unrelated question–chunk pairs: values below 0.85 often admitted chunks that were only loosely related to the question, whereas higher thresholds tended to filter out relevant chunks. A more systematic ablation of the BERTScore threshold and the choice of $n$ is left for future work.

\subsubsection{LLM for the final answer generation}
The final component in our pipeline is the LLM model that perform the summarization on the previously obtained information. The idea is that this LLM will produce better summaries if we provide it with the augmented query containing subquestions extracted from the query and the answers to those questions provided by the QA model using source document as context. The structure of the augmented prompt we send to the LLM is in Appendix A. 

In the following  section, we validate our approach. We test four LLMs and compare the summary using solely the user's query with the summary where the query is augmented with question-answer pairs.

\section{Evaluation}
\label{sec:evaluation}
For less-resourced languages, evaluation methods that rely on reference summaries are usually off the table. Therefore, we resort to methods based on automatic question generation and answering, and first describe the adaptations of QuestEval \cite{questeval} and QAGS \cite{qags}  to Slovene. Next, we describe our experimental settings and results. We end with the qualitative analysis of the results.

\subsection{QAGS}
The first step in QA evaluation framework QAGS is to extract $K$ named entities from a summary. For each extracted named entity, the QG model forms questions in such a way that the extracted named entity is a correct answer to the generated question, conditioned on the summary. Next, the QA model answers the questions generated in previous step, but it answers them conditioned on the source text. After that, another set of answers is generated by the same QA model but using the candidate summary as the only context. Finally, a similarity metric is computed between pairs of answers for every question and their average is the returned score of the metric. 

\subsection{QuestEval}
The QuestEval evaluation framework, in contrast to QAGS, generates two sets of questions instead of one. The first set of questions is created by QG model observing the generated summary solely (as is the case in QAGS), and then for that set of questions QA provides answers first conditioned on the summary, and in the next step QA model produces answers by using the source document as the only context. The second set of questions is generated based on the entire original text, and then the QA model generates answers for each question by switching between the candidate summary and the source document as the only context. The novelty in the QuestEval framework, is the introduction of the Weighter component, that performs the task of assigning weights to each question generated from the original document based on its relevance to the text. The Weighter component enables QuestEval to factor in what amount the crucial information from the source is present in the summary. Additionally, the formula for computation of the QuestEval score features the Confidence of answerability, which represents the probability of QA model not predicting a special non-answerable token as answer to a specific question. In this way, if the QA model deems important questions from the source for which the answer exists within the document as unanswerable using the candidate summary as only context, the summary will be assigned a lower score. Finally, the output of the QuestEval is the $F_1$ score, where the precision component measures consistency of the summary with the source text by comparing whether answers on the first set of questions differ when the QA model is conditioned on the summary or the original document. The recall component of QuestEval measures whether the summary contains the most important information from its source text, and is measured by comparing the answers on the second set of questions.

The definition of the QuestEval's recall component is given below. In hat formula, $W(q, D)$ is the weight assigned to question $q$ by the pre-trained Weighter component, given the source document $D$. The $\xi$ is "unanswerable" token that needs to be included in the vocabulary of QA models, and $(1 - Q_A(\xi | S, q))$ part represents the probability with which the model predicts the question is answerable. Therefore, if a question generated on the source document is assigned a high weight, but the QA model is unable to answer the question given the summary, then it would lower the QuestEval score for that summary.

\begin{align}
    Rec(D, S) = \frac{\sum\limits_{q \in Q_G(D)} W(q, D)(1 - Q_A(\xi | S, q))}{\sum\limits_{q \in Q_G(D)} W(q, D)}
\end{align}

\subsection{The experimental setting}
\label{sec:experiments}
To evaluate the effectiveness of our approach, we selected four LLMs to perform query-focused summarization. For every LLM selected as a summarizer model in our system, we utilize three different settings (see Section \ref{sec:QueryDecomposer}): 
\begin{itemize}
    \item \textbf{\texttt{decomp\_aug}} -  using another LLM to decompose the query into sub-questions that will be answered by QA component and in that way augment the prompt; 
    \item \textbf{\texttt{ner\_aug}} - NER model searches for named entities within the query, and forwards them to our QG model that creates questions using them; 
    \item \textbf{\texttt{no\_aug}} in which user's query is passed straight to the LLM as it is. 
\end{itemize}
We use the following LLMs:  GPT-4.1-mini, Gemma 2 9B-It, Llama 3.1 8B Instruct, and GaMS 9B Instruct. 

For the purpose of this evaluation, we collected a set of 21 texts in Slovene from popular Slovenian web news sources (delo.si, dnevnik.si, 24ur.com, siol.net).  The summaries generated by models for every setting are first evaluated using QAGS and QuestEval frameworks, and later manually analyzed. While the number of texts is low, it is sufficient to demonstrate the differences quantitatively and to allow for in-depth qualitative analysis.

\subsection{Quantitative evaluation}
 The results of the automatic evaluation are discussed below: for QAGS they are presented in Table \ref{tab:qags_results}, while the QuestEval scores are presented in Table \ref{tab:questeval_results}. The qualitative analysis follows in Section \ref{sec:qualitativeAnalysis}. The core of QA-based methods is to measure the similarity between answers generated using the entire original document as context and answers producing using the candidate summary as only context. In this work, the similarity between answers is measured using four different metrics: Exact match score (EM), $F_1$-score, Edit Distance and BERTScore $F_1$.

The Exact match has value 1 if two strings (answers) are identical, meaning that all words in the answer are the same and appear in identical order. If there are any differences in the arrangement of words or phrasing, the Exact match yields 0 as the score of the metric for a given answer-pair. The $F_1$ score in QA-based evaluation frameworks, is computed by measuring the number of overlapping unigrams between answer conditioned on the source document and the answer conditioned on the candidate summary. It is computed as a harmonic mean by utilizing values from precision and recall component. The Edit distance, sometimes also referred to as the Levenshtein distance measures the number of characters between two strings that need to be changed in order for strings to match. The authors of QAGS \cite{qags} reported their results using these three metrics. Finally, we decided to add BERTScore $F_1$ which is also computed by comparing those sets of answers, to allow for differently phrased answers, but are similar in meaning (at least in high-dimensional embedded space), to be assigned higher scores.

\begin{table}[htb]
\centering
\caption{Evaluation with QAGS framework across models and settings presenting the mean ± standard deviation. The best score across all models and metrics is in \textbf{bold}, and the best result per model is highlighted with \dotuline{underline dashes}.}
\label{tab:qags_results}
\renewcommand{\arraystretch}{1.2} 
\setlength{\tabcolsep}{4pt}       
\footnotesize
\begin{tabular}{lcccc}
\toprule
\textbf{Model} &   \multicolumn{4}{c}{\textbf{QAGS}} \\
\textbf{Setting} &   \textbf{$F_1$} & \textbf{EM} & \textbf{Edit Distance} & \textbf{BERTScore$_{F_1}$} \\
\midrule
\multicolumn{5}{l}{\textbf{gpt-4.1-mini}} \\
\midrule
 no\_aug    & 0.164 ± 0.307 & 0.105 ± 0.306 & 45.505 ± 24.371 & 0.685 ± 0.356 \\
 decomp\_aug  & 0.253 ± 0.389 & 0.2 ± 0.4 & \dotuline{40.436 ± 27.232} & \dotuline{0.732 ± 0.331} \\
 ner\_aug     & \dotuline{0.279 ± 0.423} & \dotuline{0.245 ± 0.43} & 42.436 ± 29.801 & 0.699 ± 0.366 \\
\midrule
\multicolumn{5}{l}{\textbf{gemma-2-9b-it}} \\
\midrule
 no\_aug      & 0.19 ± 0.349 & 0.133 ± 0.34 & 46.362 ± 25.687 & 0.614 ± 0.402 \\
 decomp\_aug  & \dotuline{0.221 ± 0.356} & \dotuline{0.152 ± 0.359} & \dotuline{41.952 ± 24.447} & \dotuline{0.756 ± 0.291} \\
 ner\_aug     & 0.219 ± 0.355 & 0.143 ± 0.35 & 44.695 ± 24.285 & 0.751 ± 0.302 \\
\midrule
\multicolumn{5}{l}{\textbf{Llama-3.1-8B-Instruct}} \\
\midrule
 no\_aug      & 0.285 ± 0.368 & 0.162 ± 0.368 & 40.048 ± 24.401 & 0.773 ± 0.298 \\
 decomp\_aug  & 0.251 ± 0.355 & 0.133 ± 0.34 & 40.695 ± 23.832 & 0.738 ± 0.33 \\
 ner\_aug     & \dotuline{\textbf{0.376 ± 0.429}} & \dotuline{\textbf{0.286 ± 0.452}} & \dotuline{36.724 ± 27.814} & \dotuline{\textbf{0.815 ± 0.261}} \\
\midrule
\multicolumn{5}{l}{\textbf{GaMS-9B-Instruct}} \\
\midrule
 no\_aug      & 0.221 ± 0.326 & 0.105 ± 0.306 & 45.933 ± 23.472 & 0.672 ± 0.369 \\
 decomp\_aug  & 0.251 ± 0.362 & 0.162 ± 0.368 & 39.886 ± 25.221 & 0.739 ± 0.319 \\
 ner\_aug     & \dotuline{0.322 ± 0.401} & \dotuline{0.209 ± 0.407} & \dotuline{\textbf{36.305 ± 27.646}} & \dotuline{0.784 ± 0.29} \\
\bottomrule
\end{tabular}
\normalsize
\end{table}

The best-performing model according to the QAGS scores is \textbf{Llama-3.1-8B-Instruct} with the \textbf{ner\_aug} setting, which uses named-entity extraction to split the query into simpler sub-questions, and it scored highest on three out of four metrics. By observing the performance indicators for this model, we can see that almost one-third of the text content for all the answer pairs matches exactly. The best result in the remaining category, Edit Distance, is achieved by \textbf{GaMS-9B-Instruct}, since for this metric the lower is better. Results clearly show that all models benefited from the information in augmented prompt retrieved by QG and QA models from source text, with the only exception occurring in case of Llama model where the model's summary achieved better results when responding to plain query than with decomp\_aug setting. A detailed inspection shows that the technique that utilizes NER for splitting query into multiple sub-units outperforms in majority of cases the approach where another LLM is used for query decomposition. This is likely the consequence of the decomposer LLM (in our experiment, GPT 4.1 mini) introducing questions of different types when breaking down the query. This forces the QA model to answer types of questions it is not trained on, which results in less significant improvement of the summary.  In the case of Gemma-2-9B model, using LLM-assisted query decomposition yielded better results according to all metrics, while for GPT-4.1 model, there is a tie between NER approach and LLM-assisted query decomposition.

Additionally, the results for EM scores reflect the property of a Bernoulli distribution in which the mean is given by $\mu = p$, where $p$ is the probability that the summary-based and source-based responses tothe  same question match, and the standard deviation is given by $\sigma = \sqrt{p(1-p)}$. Consequently, the observed standard deviations for EM are relatively large compared to the mean values, which is consistent with the discrete and high-variance nature of a binary outcome, such as word-level matching between two sets of answers.

As QAGS evaluation showed that our QFS-Composer framework results in summaries that cover more information from the source document when using an augmented prompt, we checked whether this is a consequence of LLMs producing longer summaries when the prompt is longer. We analyze the length of generated summaries in terms of character count and word count. The results are displayed in Table  \ref{tab:summary_lengths}.

\begin{table}[htb]
\centering
\caption{Character-wise and word-wise length of summaries generated by different models and settings.}
\label{tab:summary_lengths}
\renewcommand{\arraystretch}{1.2} 
\setlength{\tabcolsep}{4pt}       

\footnotesize

\begin{tabular}{lcc}
\toprule
\textbf{Model} &   \multicolumn{2}{c}{} \\
\textbf{Setting} &   Character count & Word count   \\
\midrule
\multicolumn{3}{l}{\textbf{gpt-4.1-mini}} \\
\midrule
 no\_aug    & 2080.48 ± 954.39 & 284.19 ± 126.23  \\
 decomp\_aug    & 758.22 ± 214.46 & 103.95 ± 29.38  \\
 ner\_aug    & 850.09 ± 308.53 & 117.5 ± 41.81  \\
\midrule
\multicolumn{3}{l}{\textbf{gemma-2-9b-it}} \\
\midrule
 no\_aug    & 1256.61 ± 352.64 & 177.91 ± 47.04  \\
 decomp\_aug     & 555.67 ± 215.88 & 78.14 ± 28.98  \\
 ner\_aug    & 578.33 ± 188.37 & 82.76 ± 26.23  \\
\midrule
\multicolumn{3}{l}{\textbf{Llama-3.1-8B-Instruct}} \\
\midrule
 no\_aug    & 1327.62 ± 137.52 & 196.28 ± 20.80  \\
 decomp\_aug    & 1075.19 ± 318.52 & 157.24 ± 47.76  \\
 ner\_aug    & 1157.48 ± 258.18 & 171 ± 40.04  \\
\midrule
\multicolumn{3}{l}{\textbf{GaMS-9B-Instruct}} \\
\midrule
 no\_aug    & 1016.91 ± 450.55 & 145.57 ± 63.19  \\
 decomp\_aug    & 628.47 ± 257.39 & 91.81 ± 40.41  \\
 ner\_aug    & 610.14 ± 263.57 & 89.81 ± 36.79  \\
\bottomrule
\end{tabular}

\normalsize
\end{table}

The analysis of the character- and word-level lengths of generated summaries contradicts our initial expectations. Specifically, summaries produced with augmented prompts were, on average, shorter than those generated from plain queries.

The fact that QAGS assigns higher scores to shorter summaries generated with QA-assisted prompts demonstrates that our QFS-Composer successfully produces more concise summaries with greater information density than their counterparts generated by the same LLM answering plain query. Results in Table  \ref{tab:summary_lengths} show that Llama 3.1 8B-Instruct, is producing longer summaries compared to other LLMs when augmented prompt is used. This could partially explain why the model achieved the highest scores according to QAGS, because its longer output enables it to convey more information. On the other hand, the outcome of this experiment shows a limitation of QAGS: it doesn't penalize for extra summary length or, conversely, reward shorter but information-dense summaries.

The results of evaluating summaries using the QuestEval framework are displayed in Table \ref{tab:questeval_results}. The best score is produced by the \textbf{GaMS-9B-Instruct} model in ner\_aug mode.

We got higher QuestEval scores if we let the Weighter component assign equal weights to all questions, and if the confidence of answerability is computed based on the actual output of our QA model: 0 if the output is unanswerable token or empty string, and 1 if the output is non-empty and non-whitespace string.  Additionally, due to the memory and time limitations, we have restricted the number of questions that will influence the computation of the final score in both precision and recall component of QuestEval to at most 10.
In contrast to outcomes from the QAGS evaluation, the gains in QuestEval with augmented prompt are less pronounced, except in case of \textbf{GaMS-9B-Instruct} for which a noticeable improvement happens when query is processed with the NER technique. As this is the best performing model overall, we can conclude that query augmentation is successful according to QuestEval as well.

\begin{table}[htb]
\centering
\caption{Evaluation with QuestEval framework across models and settings presenting the mean ± standard deviation. The best score across all models and metrics is in \textbf{bold}, and the best result per model is highlighted with \dotuline{underline dashes}.}
\label{tab:questeval_results}
\renewcommand{\arraystretch}{1.2}
\setlength{\tabcolsep}{4pt}
\footnotesize

\begin{tabular}{lc}
\toprule
\textbf{Model / Setting} & \textbf{QuestEval (mean ± std)} \\
\midrule
\multicolumn{2}{l}{\textbf{gpt-4.1-mini}} \\
\midrule
no\_aug      & \dotuline{ 0.351 ± 0.109 } \\
decomp\_aug  & 0.329 ± 0.113 \\
ner\_aug     & 0.332 ± 0.112 \\
\midrule
\multicolumn{2}{l}{\textbf{gemma-2-9b-it}} \\
\midrule
no\_aug      & 0.363 ± 0.087 \\
decomp\_aug  & 0.363 ± 0.099 \\
ner\_aug     & \dotuline{0.374 ± 0.126} \\
\midrule
\multicolumn{2}{l}{\textbf{Llama-3.1-8B-Instruct}} \\
\midrule
no\_aug      & 0.381 ± 0.119 \\
decomp\_aug  & 0.348 ± 0.141 \\
ner\_aug     & \dotuline{0.389 ± 0.131} \\
\midrule
\multicolumn{2}{l}{\textbf{GaMS-9B-Instruct}} \\
\midrule
no\_aug      & 0.386 ± 0.134 \\
decomp\_aug  & 0.364 ± 0.150 \\
ner\_aug     & \dotuline{\textbf{0.427 ± 0.141}} \\
\bottomrule
\end{tabular}

\normalsize
\end{table}

Relatively low scores for all models and settings can be attributed to QuestEval's dependency on exact matches between answer pairings. Often, LLMs use paraphrased expressions to describe the content of the original source, which, as a result, significantly degrades QuestEval's ability to assign correct scores to perfectly valid summaries. Moreover, the QuestEval measures the coverage of important information from the original document within the summary, and if the summary is generated with respect to a query, some information may be intentionally left out. Finally, the limit we imposed on number of questions that participate in computation of the metric together with imperfect QA and QG models contributes to the lower final score. 

\subsection{Qualitative analysis}
\label{sec:qualitativeAnalysis}
Analysis of numerical results from QAGS and QuestEval metrics is not sufficient to describe the overall behavior of the proposed QFS-Composer approach. Therefore, in this section, we take a closer look at summaries produced with \texttt{ner\_aug} technique for best-performing models \textbf{GaMS-9B-Instruct} and \textbf{Llama-3.1-8B-Instruct} and compare them to responses from the same models when there is no prompt augmentation.

Upon examining summaries, we confirm that, in general, summaries generated with an augmented prompt are shorter, but more concise and information-dense than those generated with the plain approach. Notably, for queries in which the user wants to discover all instances of a certain phenomenon, or all factors that influence a certain thing, the basic \texttt{no\_aug} approach yields a response in the form of an ordered list, while the augmented prompt approach yields plain text as output. 

We observe for \textbf{GaMS-9B-Instruct} that if a named entity is omitted from the query, it is less likely that the same entity will be mentioned in the summary. This claim is best supported by summaries for text with ID \#2, which correctly respond to the query but lack informative named entities about who and where the experiment was performed. 

A review of augmented prompts utilized by models shows the limitations of the QA model. In certain cases, the QA model couldn't find an answer to a question, although the answer definitely exists in the source text. The reason for this is that the QA model is not able to handle long texts all at once; instead, we have to split the document into chunks and rank them according to BERTScore, which is computed by comparing the chunks to the question. Hence, it is possible that a chunk not containing the answer is ranked the highest, which can lead to no output or wrong answers from the QA model. We detected that the same problem occurs in the evaluation phase. 

While manually inspecting model outputs, we found instances of LLM hallucinations. We observe that for the text in the interview form, model \textbf{GaMS-9B-Instruct} generated an off-topic response, while \textbf{Llama-3.1-8B-Instruct} hallucinated and stated that the fisherman indirectly suggested changes of the law which are completely opposite of the changes and complaints the fisherman actually talked about.

\section{Discussion and conclusion}

In our work we introduced a novel framework for query-focused summarization that we called \textbf{QFS-Composer}, and we demonstrated that by leveraging query decomposition and intra-document information retrieval using our question generation and answering models we were able to improve factual consistency and relevance of generated summaries compared to strong LLM baselines. 

Beyond experimental performance improvements, a central contribution of this work lies in the development of new assets in domain of the Slovenian NLP: QA and QG models optimized for factual supervision, translation of MOCHA dataset, and adapted metrics such as QAGS, QuestEval and RQUGE that are designed for reference-free summary evaluation. By combining these components within our framework we believe we have paved the way for more rigorous progress in summarization process for less-resourced languages. Additionally, by making these resources modular and reusable, we aim to encourage future research and downstream benchmarking.

While our results show measurable gains, several challenges remain. First, our approach relies on query decomposition and QA model accuracy, meaning that any eventual errors in these stages propagate into the final task of summarization. Therefore, the future work should focus on improving precision of each of the sub-components of QFS-Composer. Furthermore, because of hardware constraints, our QA and QG model are able to process limited amounts of texts, which consequently forced us to split original document into smaller text segments. A viable path for future research should explore whether the performance of our framework would benefit if we could enlarge the amount of context models can handle at once. Furthermore, future work should also focus on collecting human ratings summaries generated with QFS-Composer and comparing them against summaries produced directly from user queries, in order to better quantify perceived usefulness and alignment with user expectations.

By reducing the gap in linguistic resources between resource-rich and less-resourced languages, we believe this research contributes to a more inclusive NLP ecosystem, where speakers of less-resourced languages can benefit from advances in query-focused summarization technology.

\section*{Conflict of Interest Statement}

The authors declare that the research was conducted in the absence of any commercial or financial relationships that could be construed as a potential conflict of interest.

\section*{Author Contributions}
Implementation: Vuk Đuranović; Conceptualization: Marko Robnik-Šikonja; Data curation: Vuk Đuranović; Investigation: Vuk Đuranović, Marko Robnik-Šikonja; Methodology: Vuk Đuranović, Marko Robnik-Šikonja; Supervision: Marko Robnik-Šikonja; Writing original draft: Vuk Đuranović; Editing \& Text improvements: Marko Robnik-Šikonja; Vuk Đuranović; All authors approved the final submitted draft.
 
\section*{Funding}
The work was primarily funded by the Slovene Research and Innovation Agency (ARIS) through projects GC-0002, L2-50070, and the ARIS core research programme P6-0411. Additional support was provided by the EU ERA Chair grant no. 101186647 (AI4DH).


\section*{Data Availability Statement}
The datasets analyzed for this study can be found in this repository https://github.com/VukMNE/qfs-composer-paper.

\bibliographystyle{Frontiers-Harvard} 
\bibliography{bibliography}


\end{document}


\onecolumn
\firstpage{1}

\title[Supplementary Material]{{\helveticaitalic{Supplementary Material}}}

\maketitle

\section{Appendix A: Prompt templates}
\subsection{Query-decomposition prompt template}

\begin{promptblock}
Razbij dano uporabniško poizvedbo v več manjših, atomarnih podvprašanj, ...
- Vrni rezultat IZKLJUČNO kot veljaven JSON objekt z ENIM poljem:
  {{"podvprasanja": ["...", "...", "..."]}}

### Primer 1
Uporabniška poizvedba: "Generiraj povzetek o porastu primerov garij v Sloveniji in svetu."
Podvprašanja (JSON objekt):
{{"podvprasanja": [
  "Koliko primerov garij je bilo v Sloveniji lani?",
  "Koliko primerov garij so v Sloveniji obravnavali letos?",
  "Koliko ljudi z garjami je Svetovna zdravstvena organizacija ocenila leta 2017?",
  "Koliko ljudi z garjami je Svetovna zdravstvena organizacija ocenila lani?",
  "Kateri so glavni razlogi za porast garij, navedeni v besedilu?"
]}}

### Primer 2
Uporabniška poizvedba: "Generiraj povzetek o tem, kako se garje širijo in zakaj zdravljenje pogosto spodleti."
Podvprašanja (JSON objekt):
{{"podvprasanja": [
  "Kako se garje prenašajo med člani gospodinjstva?",
  "Katere navade ali napake pri terapiji prispevajo k neuspehu zdravljenja?",
  "Kdo je po besedilu najbolj izpostavljen tveganju za okužbo z garjami?",
  "Kateri simptom je tipičen za garje ponoči?"
]}}

### Primer 3
Uporabniška poizvedba: "Generiraj povzetek navodil za zdravljenje garij."
Podvprašanja (JSON objekt):
{{"podvprasanja": [
  "Kako je treba nanašati kremo na telo pri odraslih?",
  "Kako pogosto in v kakšnih intervalih je treba ponoviti nanos zdravila?",
  "Kaj je treba storiti z oblačili in posteljnino pred in po terapiji?",
  "Na katerih mestih na telesu odraslih so spremembe najpogosteje opazne?"
]}}

### Nova uporabniška poizvedba:
"{query}"

Podvprašanja (JSON objekt z enim poljem "podvprasanja"):
\end{promptblock}

\subsection{QFS augmented prompt}

\begin{promptblock}
\textbf{[USER QUERY]} 
--------------------------
Besedilo: [SOURCE DOCUMENT] 
--------------------------
V nadaljevanju so odgovori na vprašanja, povezana z poizvedbo:\\
Vprašanje: [QUESTION]
Odgovor: [ANSWER]
Vprašanje: [QUESTION]
Odgovor: [ANSWER]
...

Na podlagi zgornjih odgovorov, na kratko in jedrnato povzemite bistvo besedila v povezavi z začetno poizvedbo.
\end{promptblock}